\documentclass{article}
\usepackage{amsmath}
\usepackage{graphicx} 
\usepackage{blindtext} 
\usepackage{balance}
\usepackage[sc]{mathpazo} 
\usepackage[T1]{fontenc}  
\usepackage{microtype}    
\usepackage{adjustbox}
\usepackage[english]{babel} 

\usepackage[hmarginratio=1:1,top=32mm,columnsep=20pt]{geometry} 
\usepackage[hang, small,labelfont=bf,up,textfont=it,up]{caption} 
\usepackage{booktabs} 

\usepackage{lettrine} 

\usepackage{enumitem} 
\setlist[itemize]{noitemsep} 

\usepackage{abstract} 

\usepackage{titlesec} 
\renewcommand\thesection{\Roman{section}} 
\renewcommand\thesubsection{\roman{subsection}} 
\titleformat{\section}[block]{\large\scshape\centering}{\thesection.}{1em}{} 
\titleformat{\subsection}[block]{\large}{\thesubsection.}{1em}{} 

\usepackage{fancyhdr} 
\usepackage{titling} 

\usepackage{hyperref} 

\pretitle{\begin{center}\Huge\bfseries} 
\posttitle{\end{center}} 
\title{A Deep Learning Anomaly Detection Method in Textual Data} 
\author{%
\textsc{Amir Jafari} \\[1ex] 
}
\date{\today} 

\renewcommand{\maketitlehookd}{%

\begin{abstract}
\noindent In this article, we propose using deep learning and transformer architectures combined with classical machine learning algorithms to detect and identify text anomalies in texts. Deep learning model provides a very crucial context information about the textual data which all textual context are converted to a numerical representation. We used multiple machine learning methods such as Sentence Transformers, Auto Encoders, Logistic Regression and Distance calculation methods to predict anomalies. The method are tested on the texts data and we used syntactic data from different source injected into the original text  as anomalies or use them as target. Different methods and algorithm are explained in the field of outlier detection and the results of the best technique is presented. These results suggest that our algorithm could potentially reduce false positive rates compared with other anomaly detection methods that we are testing.
\end{abstract}
}

\begin{document}

\maketitle
\section{Introduction}

Identifying anomalous sentiment patterns, or unique textual characteristics of such patterns in a set of textual data, is known as anomaly detection in text mining. The identified anomalies could be the result of abrupt changes in decision-making for text classification problem types. If these abnormalities go unnoticed or are improperly handled, consequences could result, such as a poor performance in text classification systems \cite{wang2014anomaly}.

Anomaly, outlier or novelty detection are a complex problem in a variety of application domains where identifying outlying data is often crucial and necessary \cite{zhang2013advancements}. A pattern that is not compatible with most of the data in a dataset is named a novelty, outlier, or anomaly. An abnormal behavior can be caused for several reasons, such as data from different sources, natural variation and measurement or human errors. There is very tiny line to distinguish between novel sample or outlier/abnormal sample.

There are numerous examples in different domains and fields which they analyze the anomalies methods \cite{boukerche2020outlier}. Many web applications can benefit from the capacity to spot anomalies in streams of text data. It can be used, for instance, to identify significant events from Twitter streams, fraudulent email exchanges, and even inaccurate descriptions in maintenance logs \cite{mahapatra2012contextual}.

In \cite{survey}, authors tried to provide a solid and comprehensive overview of on anomaly detection. Existing methods in anomaly detection  and categorized them into different groups based on the techniques. For each category, they  identified key points that identify between normal and anomalous behavior.

Spectral anomaly detection techniques are one of many ways for detecting outliers. They produce a lower dimensional embeddings of the original data where anomalies and regular data are predicted to be separated from one another. The process of producing the original data after creating these lower dimensional embeddings is known as reconstruction model. In \cite{an2015variational}, they propose an anomaly detection method using variational autoencoders.

In\cite{oneclassNN}, the authors used one class neural network to detect anomalies in complex data sets. The ability of deep neural network networks is to extract rich numerical representation of data with the one-class objective of creating a tight envelope around normal data.

Detecting the abnormal behavior in dataset is a critical task in  a variety of application domains such as internet usage detection, medical diagnosis, fraudulent activities in finical domain, textual data, and patient monitoring \cite{HAUSKRECHT201347}, \cite{InternetTraffic} and \cite{Finance}.

\section{Related Work}

In Data Mining, anomaly or outlier detection can be part of the following tasks:
\begin{itemize}
  \item Classification
  \item Clustering
  \item Regression
\end{itemize}

Historically, detection of anomalies has led to explore and find of new theories and can lead into a unseen event. The outcome of anomaly detection sometimes will result in a discovery of certain pattern and facts. An outlier is an observation in a sample set of data, which deviates so much from other observations as to rise a suspicion that it was created by a different mechanism \cite{cox1984monographs}.

Assume we have a univariate normal distribution,

\begin{gather}
  f(x) = \cfrac{1}{(2 \pi \sigma ^{2})^{1/2}} e ^{-([x- \mu/\sigma]^{2})/2}\\
  (\cfrac{x-\mu}{\sigma})^{2} = (x-\mu)(\sigma^{2})^{-1}(x-\mu)
  \label{eq1}
\end{gather}

In the Eq \ref{eq1} exponent is measuring the square of deviation from mean and it is normalized by standard deviation. For $d$ dimension, the exponent is called (square of) Mahalanobis distance. The key point is that if data follows a $d$ dimensional Normal Gaussian distribution the, anomalies/outliers can be identified in the tail of the distribution. There are several weaknesses of the purely distance based approach.

\begin{itemize}
  \item Data may not follow a Normal distribution or be a mixture of distributions.
  \item Both mean and variance of exponent is $d$. For high-dimensional data this is a problem.
  \item Mean and thus variance are extremely sensitive to outliers and we are using them to and anomalies often leads to false negatives.
\end{itemize}

Another outlier detection method is principal component analysis (PCA) based technique. PCA is a dimension reduction method in which the direction of the largest eigenvalue the eigenvector projected is called the first principal component. The orthogonal direction that captures the second largest projected eigenvector is called the second principal component and so on \cite{PCANovelty}. In most cases, PCA is used for dimensionality reduction purpose however it can be used to detect outliers too.

In recent years,  deep autoencoder is used in combination with other methods. Deep autoencoder is a feedforward neural network that is trained to recreate its input \cite{AE}.

\section{Methods}

In this section, we describe the technique and methods of identifying outliers of data without labels using a deep learning autoencoder (AE) with reconstruction error. The proposed method has been divided into 2 phases.

\begin{figure}[hbt!]
\includegraphics[width=6in, height=2.5in ]{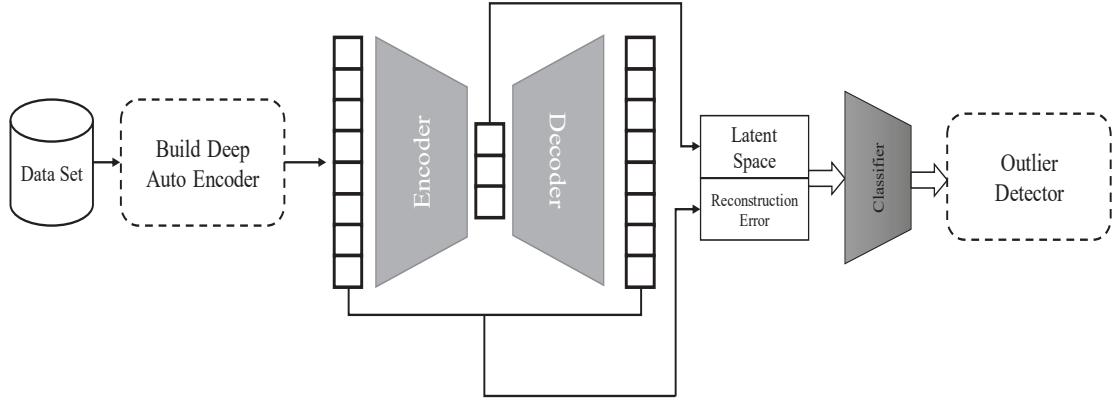}
\caption{General Architecture of Proposed Method} \label{General-Arc}
\end{figure}

The proposed approach has two basic components: dimension reduction and identification of outlier. The general scheme of the proposed method is illustrated in Figure \ref{General-Arc}

\subsection{Autoencoder}

\begin{figure}[hbt!]
\includegraphics[width=6in, height=2.5in ]{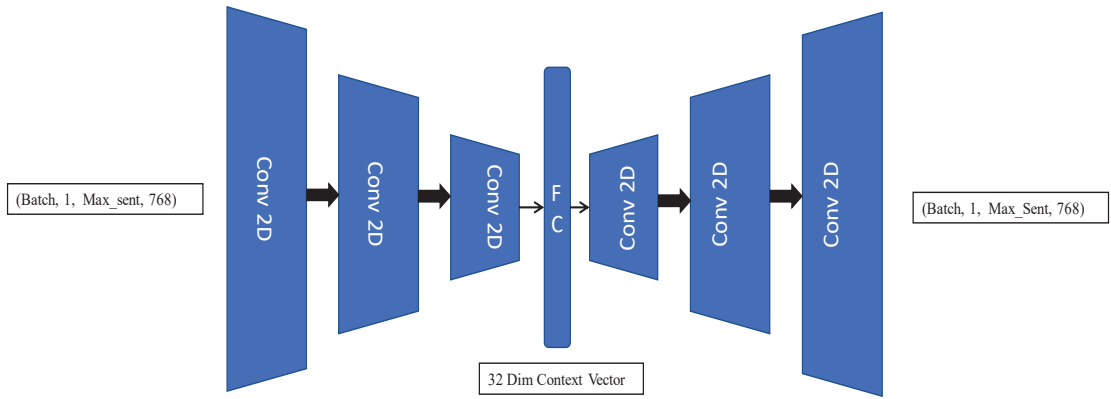}
\caption{General Architecture of Proposed Method} \label{AE}
\end{figure}

Autoencoder is a neural network architecture that accepts any inputs and try to replicate the input as an output which is the unsupervised algorithm. It learns the copy of its own input (by reducing the difference between inputs and outputs. As visualized in Figure \ref{AE}, input and output neurons are the same for AE and its hidden layer is a compressed or learned feature. In general, the AE structure is a neural network with a hidden layer at least 1 but AE is distinguished from the goal to predict the output corresponding to the specific input of NN by the purpose of reconstruct the input.

We used 2D convolutional layer in our AE architecture because the input is in 3 dimensional space.

\subsection{Sentence BERT}

Transformers have changed the landscape of natural language processing (NLP). Before transformers, recurrent neural nets (RNNs) models were used in natural machine translation and sequence prediction and text classification. The first transformer model was introduced in the 2017 paper "Attention is all you need", and become the state of art models. NLP has moved from LSTM and RNN models to transformer based models. These SOA models can solve multiple NLP tasks such as Question Answering, Text Generation , Text classification, summarization and so on.

There are significant changes in the transformer models compare to RNN type model which is in input to these layers. The richer embedding created by transformer models has more information and we get a lot enhancement in the performance due the changes.

These rich sentence embedding can be used to predict the sentence similarity such as:

\begin{itemize}
  \item Semantic Textual Similarity (STS): Identify patterns in text data.
  \item Clustering: It can be used in topic modeling.
\end{itemize}

BERT (Bidirectional Encoder Representations from Transformers) is one of the earliest and most widely used transformer models. Because transformer models focus on word or token type embedding rather than sentence type embedding, they suffer from creating sentence vector representation. Cross-encoder structure is utilized to determine the precise sentence similarity to BERT before creating the sentence transformers type model. It sends two sentences to BERT, which it then uses to output the similarity score.


Although the cross-encoder structure accurately predicts similarity scores, it is computationally expensive and cannot be scaled. Two strategies are possible in this situation. Create a prediction model first, then compute all the sentence vector representations. Second, sentence embedding can be created by averaging the values across all 512 token embedding using BERT or other transformer-based model. As a result, we can use the first [CLS] token's output for the classification task. While employing the aforementioned method speeds up prediction, it has worse accuracy than using averaged GloVe embeddings.

With the launch of sentence-BERT (SBERT) in 2019, Nils Reimers and Iryna Gurevych offer a solution to this issue of poor accuracy. For all semantic textual similarity (STS) challenges, SBERT fared better than the earlier SOTA models. It creates sentence embedding with a short inference time and is scalable. More sentence transformer models have been trained using related ideas since the SBERT paper. They have all been trained on numerous pairs of sentences, both similar and dissimilar. To provide similar embeddings for similar sentences and dissimilar embeddings, a model is optimized using a loss function like softmax loss or MSE margin loss.

\begin{figure}[hbt!]
\includegraphics[width=6in, height=2.5in ]{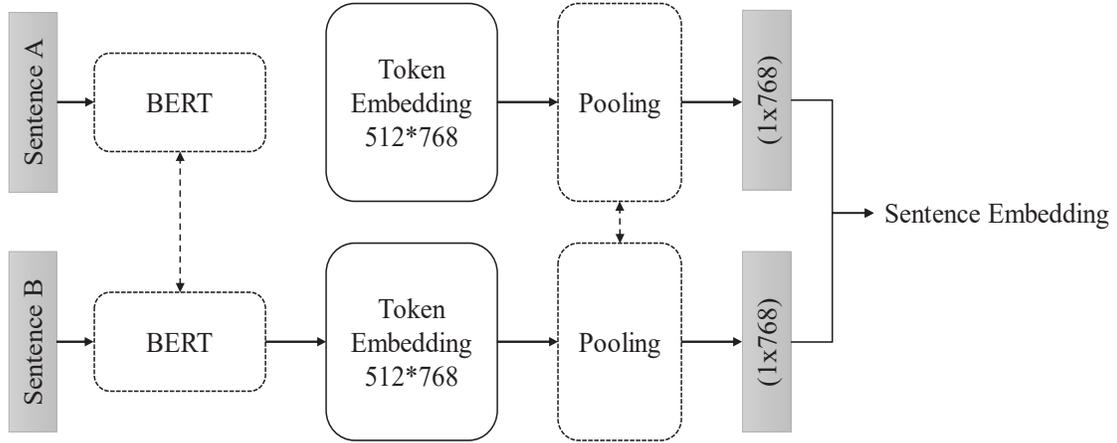}
\caption{An SBERT model applied to a sentence pair } \label{sbert}
\end{figure}

The cross-encoder architecture for sentence similarity with BERT concept is used in SBERT  but remove the classification head, and processes one sentence at a time shown in Fig \ref{sbert}. SBERT then uses mean pooling on the final output layer to produce a sentence embedding. SBERT is fine-tuned on sentence pairs, it like having two identical BERTs that share the exact same network weights.

In SBERT, the cross-encoder architecture for sentence similarity with the BERT concept is employed, but the classification head is removed, and each phrase is processed separately as illustrated in Fig \ref{sbert}. On the final output layer, SBERT then applies mean pooling to create a sentence embedding. Similar to having two identical BERTs with the same network weights, SBERT is adjusted for sentence pairings.

Sentence transformers are trained using a variety of techniques. Siamese BERT pre-training is among the most prominent and innovative techniques. The Stanford Natural Language Inference (SNLI) and Multi-Genre NLI (MNLI) corpora are employed in the "Siamese" architecture's fine-tuned use of the softmax loss function method. MNLI has 430K sentence pairings, while SNLI has 570K. Both corpora's pairings consist of a premise and a hypothesis. Each couple receives one of three labels, including:

\begin{itemize}
  \item 0 : entailment
  \item 1 : neutral
  \item 2 : contradiction
\end{itemize}

Phrase A is fed as a hypothesis into siamese BERT A and sentence B is fed as a premise into siamese BERT B using the SNLI data. The siamese BERT generates sentence embeddings using the mean, max, and [CLS]-pooling methods, respectively. Now we have two sentence embeddings. The next step is to concatenate the two and the difference is calculated to give us the element wise difference between the two vectors. Alongside the original two embeddings these are all fed into a feedforward neural net (FFNN) that has three outputs.

\section{Proposed Architecture}

\subsection{Dataset}

The goal of this work is to detect the unusual responses. The Cross-lingual Natural Language Inference (XNLI) corpus which is the extension of the Multi-Genre NLI (MultiNLI) is our core dataset. The  Stanford Sentiment Treebank (sst)is a corpus with fully labeled parse trees is used as auxiliary dataset which samples from it to be injected to XLNI dataset. A sample of 1000 text taken from sst set injected to XNLI set to create a dataset in which XNLI sample text are normal and the sst samples are outliers.

The constructed dataset (XNLI+sst) is divided to 3 sets of training, test and validation set. XNLI sample text are getting  labeled as zero target or normal sample and the injected sst data are getting label as one target or outlier. We believe outlier labeled sample are a accurate indicators to measure the performance of detection technique and find special patterns in text writings.

A preprocessed text has been used and all the text are cleaned with no special characters and tags.  Finally every samples needs to be divided into sentences in order to input it into a model.

\subsection{Model Structure}

The Auto Encoder model is trained on the all the chunked texts (sentences). The input into a AE is 3 dimensional array (1, max\_sent, feature\_dim). We set a fixed number for a max\_sent and the SBERT feature for every sentence is 768 dims. All texts in the item will be passed as input to the AE. Three convolutional layers shown in Fig \ref{AE} is built for encoder part and decoder and the fully connected layer at the last layer of the encoder is 32 dims context vector. The goal of the AE is to recreate the input and ideally if the input and out is the same then 32 latent vector can represent a text.

AE is trained on the train and ensemble split then calculate the MSE of between input and target (reconstruction error). Every texts now can be represented in 32 dimensional space. Concatenate reconstruction error to the 32 dim. That is our input to the logistic regression.

\begin{figure}[hbt!]
\includegraphics[width=6in, height=2.5in ]{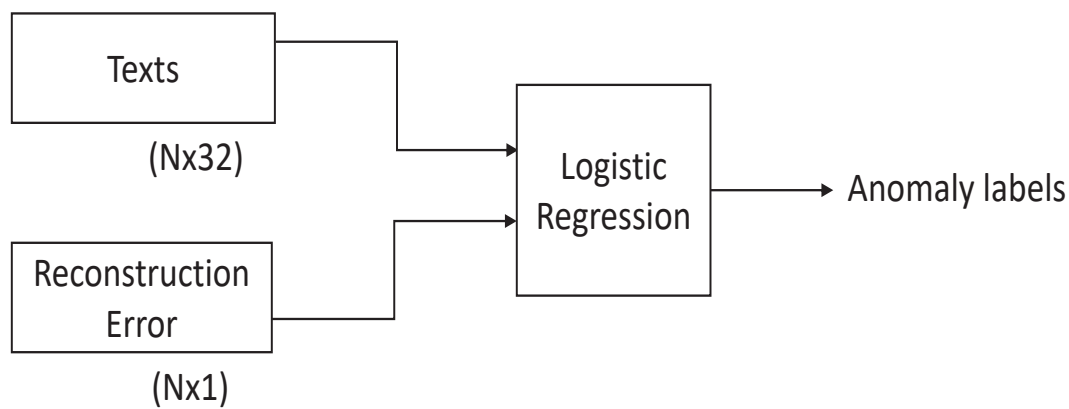}
\caption{Outlier Detector} \label{Arc}
\end{figure}

In Fig \ref{Arc} the overall architecture of the model is shown. Logistic regression is used to predict the outlier text sample labels. Since the data is so imbalanced, we used over sampling technique to avoid skewness in prediction toward the majority class.

The most commonly used metrics when evaluating anomaly detection solutions are F1, Precision and Recall. Recall is used to answer the question: What proportion of true anomalies was identified? Precision answers the question: What proportion of identified anomalies are true anomalies? F1 Score identifies the overall performance of the anomaly detection model by combining both Recall and Precision

\section{Results}

Table \ref{Outlier} shows the results of outlier detection for XNLI plus injected sst dataset. The outliers are the sst samples and the normal samples are the XNLI set. We used F1 score, precision and recall metric to evaluate the performance of model. As it is shown we have almost a very accurate model with F1 score of 86 percent.

\begin{table}[!ht]
    \centering
    \caption{Outlier Detection Results for XNLI +SST Dataset}
    \begin{tabular}{|l|l|l|l|l|l|}
    \hline
        Item & Validation Sample & F1 & Precision Score & Recall Score & Confusion Matrix \\ \hline
        XNLI +SST & 3490 & 0.86440678 & 0.918 & 0.816725979& $ \left[ \begin{array}{cc} 2284 & 82 \\ 206 & 918 \end{array}\right]$  \\ \hline
    \end{tabular}
    \label{Outlier}
\end{table}

Confusion matrix results is also presented in the table. 82 sample text are normal text misclassified as outlier and 206 samples misclassified are normal which is outlier.


\section{Conclusion and Future Work}

The deep learning approach in outlier detection or anomaly detection would be a great filter for text classification engines. This work shows the outlier detector which uses a transformer based model adds more context values into anomaly detection models which enhanced the classical distance based techniques. The novel approach of using sentence transformer to convert the texts to a rich numerical representation of whole text and use the Convolutional Auto Encoder to compress the text a vector is unique. Now we can use our machine learning tool to classify or identify the abnormal behaviors or oulierness using the very rich dense vector which represents the whole text.  For future work, we could verify the performance on the outlier detector on students Kaggle essay data set.

\bibliographystyle{IEEEtran}
\balance
\bibliography{mybib_outlier.bib}

\end{document}